\documentclass[sigconf]{acmart}

\usepackage{bm}
\usepackage{array}
\usepackage{color}
\usepackage[normalem]{ulem}
\usepackage{colortbl}
\usepackage{textcomp}
\usepackage{stfloats}
\usepackage{hyperref}
\usepackage{verbatim}
\usepackage{graphicx}
\usepackage{booktabs}
\usepackage{subfig}
\usepackage{paralist}
\usepackage{colortbl}

\allowdisplaybreaks 

\usepackage{multirow} 
\usepackage{booktabs} 
\usepackage{arydshln} 

\AtBeginDocument{%
  \providecommand\BibTeX{{%
    \normalfont B\kern-0.5em{\scshape i\kern-0.25em b}\kern-0.8em\TeX}}}

\newcommand{\paratitle}[1]{\vspace{1.4ex}\noindent \textbf{#1}}

\newcommand{\mfnt}[1]{\fontsize{6}{9}\selectfont #1}

\copyrightyear{2023}
\acmYear{2023}
\setcopyright{acmlicensed}
\acmConference[MM '23] {Proceedings of the 31st ACM International Conference on Multimedia}{October 29--November 3, 2023}{Ottawa, ON, Canada.}
\acmBooktitle{Proceedings of the 31st ACM International Conference on Multimedia (MM '23), October 29--November 3, 2023, Ottawa, ON, Canada}
\acmPrice{15.00}
\acmISBN{979-8-4007-0108-5/23/10}
\acmDOI{10.1145/3581783.3612053}

\begin{document}

\title[Revisiting Conversational Emotion Recognition]{Revisiting Disentanglement and Fusion on Modality and Context in Conversational Multimodal Emotion Recognition}

\author{Bobo Li}
\orcid{0000-0002-0513-5540}
\affiliation{%
    \institution{Key Laboratory of Aerospace Information Security and Trusted Computing, Ministry of Education, School of Cyber Science and Engineering, Wuhan University}
  \country{}
}
\email{boboli@whu.edu.cn}

\author{Hao Fei}
\affiliation{%
  \institution{NExT Research Center, School of Computing, National University of Singapore}
  \country{}
  }
\email{haofei37@nus.edu.sg}

\author{Lizi Liao}
\affiliation{%
  \institution{School of Computing and Information Systems, Singapore Management University}
  \country{}}
\email{lzliao@smu.edu.sg}

\author{Yu Zhao}
\affiliation{%
  \institution{Tianjin University}
  \country{}}
\email{zhaoyucs@tju.edu.cn}

\author{Chong Teng}
\affiliation{%
  \institution{Wuhan University}
  \country{}}
\email{tengchong@whu.edu.cn}

\author{Tat-Seng Chua}
\affiliation{%
  \institution{National University of Singapore}
  \country{}}
\email{dcscts@nus.edu.sg}

\author{Donghong Ji}
\affiliation{%
  \institution{Key Laboratory of Aerospace Information Security and Trusted Computing, Ministry of Education, School of Cyber Science and Engineering, Wuhan University}
  \country{}}
\email{dhji@whu.edu.cn}

\author{Fei Li}
\authornote{Fei Li is the corresponding author.}
\affiliation{%
  \institution{Key Laboratory of Aerospace Information Security and Trusted Computing, Ministry of Education, School of Cyber Science and Engineering, Wuhan University}
  \country{}}
\email{lifei_csnlp@whu.edu.cn}

\renewcommand{\shortauthors}{Bobo Li et al.}

\begin{abstract}
It has been a hot research topic to enable machines to understand human emotions in multimodal contexts under dialogue scenarios, which is tasked with multimodal emotion analysis in conversation (MM-ERC).
MM-ERC has received consistent attention in recent years, where a diverse range of methods has been proposed for securing better task performance.
Most existing works treat MM-ERC as a standard multimodal classification problem and perform multimodal feature disentanglement and fusion for maximizing feature utility.
Yet after revisiting the characteristic of MM-ERC, we argue that both the \emph{feature multimodality} and \emph{conversational contextualization} should be properly modeled simultaneously during the feature disentanglement and fusion steps.
In this work, we target further pushing the task performance by taking full consideration of the above insights.
On the one hand, during feature disentanglement, based on the contrastive learning technique, we devise a Dual-level Disentanglement Mechanism (DDM) to decouple the features into both the modality space and utterance space.
On the other hand, during the feature fusion stage, we propose a Contribution-aware Fusion Mechanism (CFM) and a Context Refusion Mechanism (CRM) for multimodal and context integration, respectively.
They together schedule the proper integrations of multimodal and context features. 
Specifically, CFM explicitly manages the multimodal feature contributions dynamically, while CRM flexibly coordinates the introduction of dialogue contexts.
On two public MM-ERC datasets, our system achieves new state-of-the-art performance consistently. 
Further analyses demonstrate that all our proposed mechanisms greatly facilitate the MM-ERC task by making full use of the multimodal and context features adaptively.
Note that our proposed methods have the great potential to facilitate a broader range of other conversational multimodal tasks.
\end{abstract}


\maketitle

\section{Introduction}

\begin{figure}[!t]
\includegraphics[width=0.98\columnwidth]{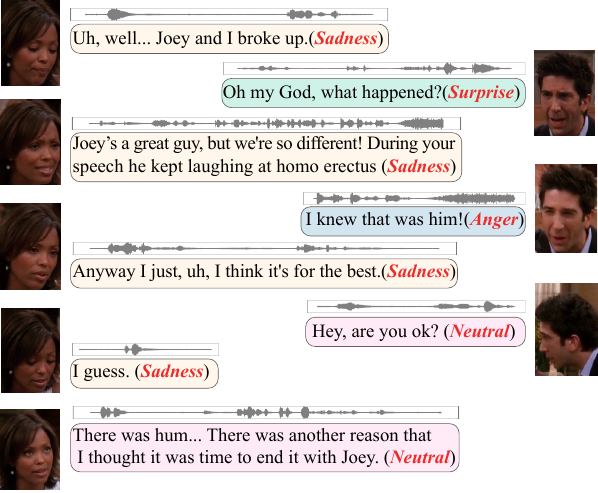}
  \vspace{-1mm}
\caption{An example of multimodal conversation from the MELD dataset~\cite{poriaHMNCM-acl-2019}.
Each utterance comes with three modalities of content: video, audio, and text.
The goal of MM-ERC is to recognize the emotion label 
of each utterance.
}
\label{fig:example}
  \vspace{-5mm}
\end{figure}

The analysis of conversational emotions~\cite{poriaHMNCM-acl-2019} has received growing attention and has been applied in various downstream tasks, like empathetic response generation~\cite{welals-emnlp-2021,gaoier-emnlp-2021,hutae-tac-2023} and mental disease treatment~\cite{sahtme-sigir-2022}.
Recently, the research on conversational emotion analysis has extended the focus from text to multiple modalities such as video and audio~\cite{poriaHMNCM-acl-2019, ghosal-cosmic-2020, humm-acl-2021}.
As illustrated in Figure~\ref{fig:example}, multimodal emotion recognition in conversation (named MM-ERC) aims to detect the emotion label for each utterance in a given dialogue by jointly considering auditory, visual, and textual content.
The introduction of audio and video compensates for the limitation of solely depending on text features and thus enriches the features used for emotion recognition.

A good number of efforts have been devoted to building effective MM-ERC models and secured promising performance, where the core idea is to effectively disentangle different modalities and then properly fuse them so as to maximize the efficacy of multimodal features for the task \cite{zadehtf-emnlp-2017,humm-acl-2021,chelw-mm-2021,hutum-emnlp-2022,humfn-icassp-2022}.
However, MM-ERC intrinsically involves two simultaneous key ingredients: \emph{multiple feature modality} and \emph{conversational contextualization}.
While the majority of existing models treat MM-ERC as a typical multimodal classification problem, focusing predominantly on either multimodality or context modeling, the relationship between dialogue context and multimodal feature consistency is often neglected.
By revisiting the task of MM-ERC, we note that a sound and effective MM-ERC system should place proper attention to simultaneously modeling the multimodality and contextualization during the feature disentanglement step and fusion step.

\begin{figure*}[!t]
\includegraphics[width=2.0\columnwidth]{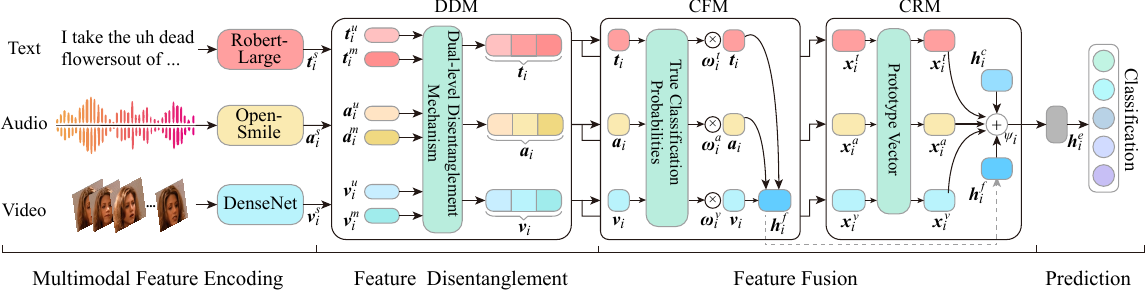}
\vspace{-1mm}
\caption{Overall framework of the proposed DF-ERC model.
DDM: Dual-level disentanglement Mechanism;
CFM: Contribution-aware Fusion Mechanism;
CRM: Context Refusion Mechanism.
}
\label{fig:main_model}
\vspace{-3mm}
\end{figure*}

\paratitle{Feature Disentanglement.} 
The purpose of feature disentanglement is to extract the critical features from the original feature spaces and weaken the influence of irrelevant features, since multimodal inputs often contain features unrelated to emotion recognition (e.g., background video and noisy audio).
While existing models, such as MISA~\cite{hazarika-mm-2020} and FDMER~\cite{yang-mm-2022}, propose sophisticated disentanglement mechanisms for single pieces of utterance, disentangling on the conversational contexts has not been considered.
On the one hand, different modality features within one utterance should exhibit similarities because, intuitively, multimodal signals under the same utterance can be semantically consistent in representing an identical emotion.
On the other hand, features from different utterances with the same modality share similarities in modality-specific characteristics (e.g., timbre, facial expression, and strong wording), which may seem trivial for other modalities but are useful for recognizing emotions in the specific modality space.
Feature disentanglement without effectively considering both the modality level and utterance level will inevitably limit further performance improvement of MM-ERC.
Unfortunately, to the best of our knowledge, no existing research explores the disentanglement under these two aspects, indicating a potential research gap.

\paratitle{Feature Fusion.} 
The disentangled features from the above step further need to be properly fused, during which reasonable weights are assigned to maximize the utility of features for emotion prediction.
Since different clues in varied modalities serve distinct contributions to the final prediction, fusing features across modalities has been extensively considered in existing MM-ERC studies \cite{chelw-mm-2021, humm-acl-2021, humfn-icassp-2022, chumfn-cvprw-2022}, with many sophisticated methods, such as tensor fusion~\cite{zadehtf-emnlp-2017},  graph convolutional networks~\cite{humm-acl-2021}, gating mechanisms~\cite{humfn-icassp-2022}.
However, no controllable weights were utilized in previous works, which may risk one modality dominating the multimodal fusion process~\cite{penbml-cvpr-2022} and potentially limiting the overall performance.
Yet we note that the utterance-level fusion should also receive sufficient attention.
Intuitively, it is less necessary to further introduce moderate history utterance contexts for prediction when the multimodal signals within the current utterance have indicated a clear emotion tendency in high consistency.
Instead, aggressively feeding all the historical contexts would rather deteriorate the inference.
For example, in Figure~\ref{fig:example}, fully considering all previous dialogue contexts might lead to an incorrect emotion determination for the last utterance as ``Sadness''.
This could happen due to the negative neighbor context (i.e., the emotion of the second-last utterance being "Sadness") and the negative atmosphere conveyed throughout the dialogue.
Therefore, properly fusing the features from both multimodal ones and dialogue contexts is non-trivial.

In light of the above observations, in this work, we develop a niche targeting solution, i.e., \textbf{DF-ERC} (\textbf{D}isentanglement \& \textbf{F}usion for \textbf{E}motion \textbf{R}ecognition in \textbf{C}onversation), to fill the gaps and help achieve higher performance in MM-ERC.
As shown in Figure~\ref{fig:main_model}, our system comprises four tiers.
First, the raw multimodal inputs of dialogues are encoded into various feature extractors to obtain corresponding features.
Then, the feature disentanglement layer performs feature disentanglement, where a \textbf{Dual-level Disentanglement Mechanism} (DDM) is proposed. 
DDM employs contrastive learning~\cite{tiascl-emnlp-2021} to push the feature vectors of different modalities or different utterances away, thereby disentangling features at the modality level and utterance level, respectively.
Next, the feature fusion layer performs modality-level and context-level integration, in which we propose a \textbf{Contribution-aware Fusion Mechanism} (CFM) and a \textbf{Context Refusion Mechanism} (CRM) for multimodal and context fusion, respectively.
CFM fuses multimodal features based on the true classification probabilities~\cite{corbiafp-nips-2019} of each modality as their contributions, where such dynamic weighting advances in more controllable feature coordination.
In contrast, CRM flexibly schedules the introduction of historical dialogue contexts into the current utterance via a novel emotion-prototype learning strategy.
Specifically, CRM calculates the consistency degree of all modality features within an utterance, where a lower consistency degree triggers the model to bring in more contexts for reassurance.
Finally, the fused overall multimodal and contextual features are used for the emotion label prediction.

To evaluate the efficacy of our proposed approach, we performed extensive experiments on two widely-used benchmarks, namely MELD~\cite{poriaHMNCM-acl-2019} and IEMOCAP~\cite{busied-lre-2007}.
DF-ERC achieves state-of-the-art performance on overall results and most of the fine-grained emotion categories, demonstrating its effectiveness and stability.
Furthermore, we find that DDM was able to effectively disentangle the features of different modalities or utterances (see Figure~\ref{fig:exp_cl}), 
and the disentangled features played a crucial role in the process of feature fusion (see Figure~\ref{fig:uddm} and Figure~\ref{fig:mddm}).
Additionally, both CFM and CRM play vital roles, as demonstrated by ablation studies (see Table~\ref{tab:ablation0}) and in-depth analysis (see Section~\ref{sec:indep}).
Especially noteworthy is that CRM outperforms the models with no context or full context engagement, demonstrating its superiority (see Figure~\ref{fig:exp_sim}). 

Overall, our contributions are four-fold:
\begin{compactitem}

\item We revisit the MM-ERC task and, for the first time, propose DF-ERC to enhance the task by performing disentanglement and fusion under both the modality and context perspectives.

\item Technically, we propose three novel and effective mechanisms to disentangle and fuse both multimodal and contextual features.

\item Empirically, our system achieves state-of-the-art 
performance on two benchmarks.
\item Our proposed methods have great potential for facilitating a broader range of conversational multimodal applications.
\end{compactitem}

\section{Related Work}
Multimodal sentiment analysis~\cite{morency-icmi-2011, wllmer-iis-2013, ringkimsa-acm-2023} aims to extract sentiments or emotions using multiple modality resources, such as text (transcripts), acoustic (audio) and visual modalities. 
However, discrepancies across different modalities pose a challenge to the model. 
To address this issue, some studies have focused on modality alignment~\cite{tsai-acl-2019} and minimizing the discrepancies between modalities~\cite{yulms-aaai-2021,mai-aaai-2020}. 
Moreover, the style of modality fusion can impact the model performance, leading to the exploration of effective fusion methods, such as hierarchical mutual information~\cite{hanimf-emnlp-2021, majumder-kbs-2018}, reconstruct loss~\cite{hazarika-mm-2020}, and graph neural network~\cite{bagher-acl-2018, yangmsd-acl-2021, huangtgc-icmi-2021}. 
Additionally, leveraging contextual information to predict dynamic emotions is also a popular approach~\cite{akhtarmlm-naacl-2019, chauhancia-acl-2019, chencfm-aaai-2020, ghosalcia-emnlp-2018}. 
However, the use of controllable weights to fuse multimodal features has not been considered in any of the existing approaches, which can limit their performance in practice and is one of the main focuses of our study. 

Emotion Recognition in Conversation (ERC) \cite{lee-interspeech-2009, mairesse-jair-2007, pereira-survey-2022} is a subfield of affective computing that aims to recognize emotions for each utterance within a conversation. 
To develop the model, some studies focus on leveraging dialogue-related features, such as speaker-oriented dialogue modeling\cite{majaar-aaai-2019, ghosal-dgcn-2019, hu-dcrn-2021}, context-aware modeling~\cite{schuller-tac-2010, zhaomcd-coling-2022}, hierarchical feature modeling~\cite{lihitrans-coling-2020, lihi-corr-2020, leechoi-emnlp-2021}, and emotion transition~\cite{song-flow-2022, bansem-mmmpie-2022}. 
With the development of multimodal technology~\cite{zhangmlf-tmm-2015, tadmml-pami-2017, summarm-tomm-2023}, the research scope of ERC has been extended to multimodal scenarios. 
Many studies have explored multimodal fusion methods for the MM-ERC task, such as multimodal dynamic fusion~\cite{humfn-icassp-2022, liacct-aslp-2021}, hierarchical fusion~\cite{chumfn-cvprw-2022}, and adaptive modality drop~\cite{chelw-mm-2021}. 
Although existing adaptive methods have been proposed, they neglect some crucial aspects, such as the contribution of each utterance and the relationship between modality consistency and the involvement of context, which are the main focuses of our paper.

\section{Framework}

Given a dialogue $D$ = $\{u_0, u_1,\cdots, u_n\}$, where $u_i$ represents an utterance, the MM-ERC task aims to recognize the emotion type $e_i$ corresponding to each utterance $u_i$.
In each $u_i$, there are three kinds of data, namely text, audio, and video, which are used to predict $e_i$.
$e_i$ belongs to a pre-defined set of emotion labels, such as \textit{angry}, \textit{sadness}, \textit{joy}, etc.
To approach the task, we introduce a novel framework, termed DF-ERC, illustrated in Figure~\ref{fig:main_model}, which performs four tiers of propagation for emotion prediction.
Subsequently, we elaborate on the specific techniques employed at each step.

\subsection{Multimodal Feature Encoding}

Given a dialogue $D$, we first perform feature extraction for each utterance $u_i$ simultaneously. 
In this paper, we follow up-to-date previous works~\cite{song-flow-2022,chumfn-cvprw-2022} and employ RoBerta~\cite{liurop-arxiv-2019} to obtain contextualized text features. 
Specifically, all the utterances are concatenated and fed into a pre-trained language model (PLM) following the way in Span-BERT~\cite{mansip-tacl-2020}.
The dialogue is represented as 
\begin{gather}
\mathcal{D}= [\text{[CLS]},w_{11},w_{12},\cdots, w_{1l_1},w_{21},\cdots,w_{nl_n},\text{[SEP]}], \notag \\
\bm{H} = \text{PLM}(\mathcal{D}), \\ 
\bm{t}^s_i = \text{MeanPooling}(\bm{H}[start_i,end_i]), \notag
\end{gather}
where $w_{ij}$ is the $j$-th token in $u_i$ and $l_i$ is the length of $u_i$, $start_i$ and $end_i$ are the indices of the head and tail tokens of $u_i$ in the sequence $\mathcal{D}$, and $\bm{t}^s_i$ is the text feature for utterance 
$u_i$.

For audio and visual content, following the approach described in previous work~\cite{humm-acl-2021, humfn-icassp-2022}, we adopt OpenSmile~\cite{schurre-sc-2011} and DenseNet~\cite{huangdcc-cvpr-2017} pre-trained on the Facial Expression Recognition Plus (FER+) corpus~\cite{barsoum-icmi-2016} as feature extractors.
Finally, we obtain an audio feature $\bm{a}^s_i$ and a visual feature $\bm{v}^s_i$ for each utterance 
$u_i$.

\subsection{Dual-level Disentanglement Mechanism (DDM)}

\begin{figure}[!t]
\includegraphics[width=0.98\columnwidth]{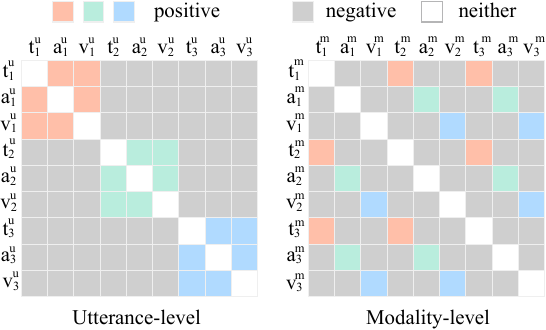}
\caption{
Dual-level disentanglement mechanism, where $\bm{t}^*_i$, $\bm{a}^*_i$, and $\bm{v}^*_i$ denote text, audio, and video features, respectively.
As seen, utterance-level disentanglement pulls the features of the same utterance close and pushes away the features of different utterances, while modality-level disentanglement pulls the features of the same modality close and pushes away the features of different modalities.
}
\label{fig:cl_matrix}
\vspace{-4mm}
\end{figure}

It should be noted that directly utilizing raw multimodal features for emotion analysis is problematic because they are entangled and noisy due to the unconstrained extraction process. 
Thus, it is necessary to disentangle multimodal features in order to refine them and boost the performance of downstream tasks.
In this paper, we propose a dual-level disentanglement mechanism to disentangle raw features in both utterance and modality levels, 
as illustrated in Figure~\ref{fig:cl_matrix}.
At the modality level, we apply an MLP layer to the features of different modalities to derive modality-level representations:
\begin{gather}
\bm{t}^m_i / \bm{a}^m_i / \bm{v}^m_i = \text{MLP}^m_{t/a/v}(\bm{t}^s_i/\bm{a}^s_i / \bm{v}^s_i).
\label{eq:tmi}
\end{gather}
Next, a list $\bm{R}^m$, containing the items $\bm{t}^m_i, \bm{a}^m_i, \bm{v}^m_i$ ($i\in[1,n]$), is created as follows: $\bm{R}^m = [\bm{t}^m_1, \bm{a}^m_1, \bm{v}^m_1, \bm{t}^m_2, ..., \bm{t}^m_n, \bm{a}^m_n, \bm{v}^m_n]$.
We then apply contrastive learning to these features in order to draw features of the same modality closer to each other and push features of different modalities away, formalized as below:
\begin{gather}
\mathcal{L}_{cl}^{m} = -\sum_{i}^{n}\sum_{\bm{h}_k\in \bm{R}^m_{i+}} \log \frac{e^{sim(\bm{h}_i,\bm{h}_k) / \tau }}{\sum_{j!=i}^{3n}e^{\mathrm{sim}(\bm{h}_i,\bm{h}_j)/\tau }},
\label{eq:clm}
\end{gather}
where $\bm{h}_i$ is the $i$-th item in $\bm{R}^m$.
Note that $\bm{h}_k \in \bm{R}^m_{i+} = \{\bm{x}_j|j\equiv\ i (\mathrm{mod}\ 3),1< j\leq 3n\}$ has the same modality as $\bm{h}_i$, which can be considered as positive instances.
Here $\tau$ is a temperature parameter, and $sim(\bm{h}_i, \bm{h}_k)$ denotes the cosine similarity between two vectors.

At the utterance level, contrastive learning is exploited in a similar manner, clustering the multimodal features of the same utterance and disentangling the features of different utterances, formulated as below: 
\begin{gather}
\bm{t}^u_i / \bm{a}^u_i / \bm{v}^u_i = \text{MLP}^u_{t/a/v}(\bm{t}^s_i/\bm{a}^s_i / \bm{v}^s_i), 
\label{eq:tui}
\\
\mathcal{L}_{cl}^{u} = -\sum_{i}^{n}\sum_{\bm{h}_k\in \bm{R}^u_{i+}} \log \frac{e^{sim(\bm{h}_i,\bm{h}_k) / \tau }}{\sum_{j!=i}^{3n}e^{\mathrm{sim}(\bm{h}_i,\bm{h}_j)/\tau }},
\label{eq:clu}
\end{gather}
where $\bm{h}_i \in \bm{R}^u=[\bm{t}^u_1, \bm{a}^u_1, \bm{v}^u_1,\bm{t}^u_2,\cdots, \bm{t}^u_n,\bm{a}^u_n,\bm{v}^u_n]$, and $\bm{h}_k \in \bm{R}_{i+}^u = \{ \bm{x}_j  | \lfloor i / 3 \rfloor = \lfloor j / 3 \rfloor, 0< j \leq 3n \}$ denotes the feature in the same utterance with $\bm{h}_i$.
$\lfloor x \rfloor $ is the round down symbol.
Finally, we use residual connections to concatenate raw features with disentangle features, and two loss functions for contrastive learning are also combined: 
 \begin{align}
\bm{t}_i &= [\bm{t}^s_i; \bm{t}^m_i; \bm{t}^u_i], \label{eq:tfi}\\
\bm{a}_i &= [\bm{a}^s_i; \bm{a}^m_i; \bm{a}^u_i],\label{eq:afi} \\
\bm{v}_i &= [\bm{v}^s_i; \bm{v}^m_i; \bm{v}^u_i], \label{eq:vfi}\\
\mathcal{L}_{cl} &= \mathcal{L}_{cl}^m + \mathcal{L}_{cl}^u.
\label{eqn:ddm_output}
\vspace{-1pt}
 \end{align}

\subsection{Contribution-aware Fusion Mechanism (CFM)}

\begin{figure}[!t]
\includegraphics[width=0.98\columnwidth]{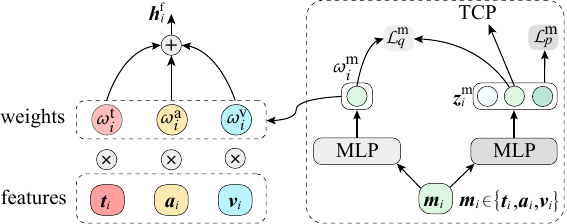}
\vspace{-1mm}
\caption{
Contribution-aware fusion mechanism. $\bm{t}_i$, $\bm{a}_i$ and $\bm{v}_i$ are the features from DDM (see Eqs.~\eqref{eq:tfi} to \eqref{eq:vfi}).
$\omega_i^t$, $\omega_i^a$ and $\omega_i^v$ are the contributions of different modalities (see Eq.~\eqref{eq:omega}), which are given by three contribution prediction networks $\mathrm{MLP}_m^p$ ($m\in\{t,a,v\}$).
These networks are trained as students based on true classification probabilities (TCPs) given by three teacher networks $\mathrm{MLP}_m^q$.
}
\label{fig:confident}
\vspace{-3mm}
\end{figure}

\vspace{-0.5mm}

As different modalities have different importance for the final emotion label prediction, they should be assigned different fusion weights in the modality fusion process.
Here we adopt a contribute-aware adaptive fusion module to assign the weight of each modality, which can give a dynamic weight according to their prediction performance, as illustrated in Figure~\ref{fig:confident}.
Specifically, we apply a classifier on the representation of each modality and obtain the true classification probability (TCP)~\cite{corbiafp-nips-2019, hanmdd-cvpr-2022} as their contribution in the fusion process,
which can be obtained via:
\begin{gather}
\bm{z}_i^{t/a/v} = \text{Softmax}(\text{MLP}^q_{t/a/v}(\bm{t}_i/ \bm{a}_i/ \bm{v}_i)), \\
\text{TCP}_i^m = (\bm{z}_i^m)_{I^*_i},m\in\{t,a,v\},
\end{gather}
where $\bm{z}_i^m$ is the prediction probability, and $I^*_i$  is the index of golden emotion label for $u_i$.
Obviously, $\rm{TCP}_i^m \in (0,1)$ denotes how likely the prediction result is right.
A larger TCP value indicates the feature representation $\bm{t}_i / \bm{a}_i / \bm{v}_i$ can yield a correct prediction result and verse visa.
Therefore, we plan to adopt TCP to represent the fusion weight for each modality.
However, a significant challenge arises during the evaluation phase as the true emotion label is unknown, making it impossible to directly utilize TCP as the weight.
To keep the consistency of training and test processes, we adopt a predicted value, which is trained to be close to TCP, as the weight of each modality:
\begin{gather}
\omega_i^{t/a/v} = \text{Sigmoid}(\text{MLP}^p_{t/a/v}(\bm{t}_i/\bm{a}_i/\bm{v}_i)).
\label{eq:omega}
\end{gather}
To achieve the goal that we mentioned before,
the following loss functions are used:
\begin{align}
\mathcal{L}_p^m &= -\sum_{ i=1 } ^n { \log(\bm{z} _i ^m (I^*_i))},\\
\mathcal{L}_q^m &= \sum_{ i=1 } ^n { \text { MSE } (\rm{TCP}_i^m, \omega_i^m)}, \label{eq:mseomega}\\
\mathcal{L}_{con} &= \sum _{m\in\{t,a,v\}} { (\mathcal{L} _p ^m + \mathcal{L} _q ^m)},
\end{align}
where, $\mathcal{L}_p^m$ represents the prediction label loss and $\mathcal{L}_q^m$ is the prediction TCP loss.
Finally, the fused multimodal features can be formulated as:
\begin{gather}
\bm{h}^f_i = \omega_i^t \bm{t}_i + \omega_i^a \bm{a}_i + \omega_i^v \bm{v}_i.
\end{gather}

\subsection{Context Refusion Mechanism (CRM)}
\begin{figure}[!t]
\includegraphics[width=0.98\columnwidth]{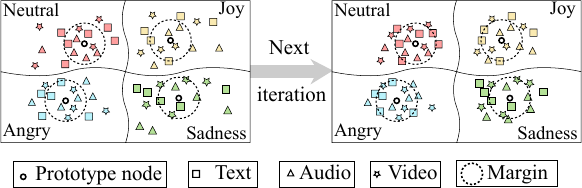}
\caption{
An illustration of the process of prototype-based alignment module.
}
\label{fig:prototype}
\vspace{-4mm}
\end{figure}

Except for fusing multimodal features, contextual feature fusion is also important, especially when 
multimodal features contradict each other regarding emotion prediction.
Therefore, we compute the agreements among multimodal features as the weights to determine how many contextual features should be incorporated.
However, since multimodal features are not aligned according to emotions,
it may be inaccurate to directly compute the similarity based on multimodal features.
To solve this problem, we propose a prototype-based alignment module (as shown in Figure~\ref{fig:prototype}) to learn the emotion-specific representations of multimodal features.
Specifically, in each training epoch, we maintain a prototype vector for each kind of emotion:
\begin{align}
\bm{x}_i^{t/a/v} &= \text{MLP}^r_{t/a/v}(\bm{t}_i / \bm{a}_i / \bm{v}_i), \\
R_r^k &= \frac{1}{|N_r^k|}(R_{r-1}^k \cdot N_{r-1}^k + \sum_{m\in\{t,a,v\}}\sum_{\text{I}^*_i = k}{\bm{x}_i^m}),
\vspace{-2pt}
\end{align}
where $R_t^k$ represents the prototype of the $k$-th emotion in the $t$-th epoch, $N_r^k=N_{r-1}^k + 3\cdot \sum_i^n 1_{I^*_i==k}$ is the size of the $k$-th emotion in the $r$-th round.
The prototype vector is updated in each iteration based on the previous values and the multimodal features in the current epoch. 
To ensure that each feature is close to its prototype vector, we use a margin-based loss function based on the mean squared error (MSE):
\begin{gather}
\mathcal{L}_{sim} = \frac{1}{3n}\sum_{i=1}^n\sum_{m,k}\sum_{I^*_i==k} max(\text{MSE}(R_r^k, \bm{x}_i^m) - \beta, 0),
\vspace{-1pt}
\end{gather}
where $\mathcal{L}_{sim}$ denotes the loss function and $\beta$ represents the margin. 
If the MSE between a feature vector and its corresponding prototype vector is less than the margin, the model will not be updated.
In other words, the feature vector is expected to be close to the prototype vector but not necessarily identical to it, in order to avoid all feature vectors becoming indistinguishable.
Once the multimodal feature vectors are aligned, the comprehensive similarity between different modalities in the utterance $u_i$ can be computed as:
\begin{gather}
\psi_i = \frac{1}{3}(\cos(\bm{x}^t_i,\bm{x}^a_i) +\cos(\bm{x}^t_i,\bm{x}^v_i) + \cos(\bm{x}^a_i,\bm{x}^v_i)
).
\label{eq:phi}
\end{gather}

In the context fusion stage, we first utilize a bidirectional long short-term memory (Bi-LSTM) to generate contextual representations of the utterances as follows:
\begin{gather}
 \bm{h}^c_1, \bm{h}^c_2,\cdots, \bm{h}^c_n = \mathrm{BiLSTM}([\bm{h}_1^f,\bm{h}_2^f,\cdots,\bm{h}_n^f]).
 \label{eq:hei}
\end{gather}
We then concatenate fused features with context-aware features as follows\footnote{
We also conducted an experiment by directly multiplying $(1-\psi_i)$ with each $\bm{h}_*^f$ in Eq.~\eqref{eq:hei} instead of $\bm{h}_i^c$ in Eq.~\eqref{eqn:context_feature}, and obtained a similar result. Therefore, we adopted the more concise fusion approach as shown in Eq.~\eqref{eqn:context_feature}.}:
\begin{gather}
\bm{h}^e_i = [\bm{h}_i^f, \bm{h}_i^c \cdot (1 - \psi_i)],
\label{eqn:context_feature}
\end{gather}
where $\bm{h}^e_i$ represents the final representation of each utterance that is fused with multimodal and contextual information. 

\begin{table*}[!t]
    \centering
    \renewcommand*{\arraystretch}{1.25}
    \caption{
Comparisons with the baselines. 
`W-F1' refers to weighted F1 scores.
The results with \uwave{waveline} denote the best baseline results.
The results with $^\star$ denote significance at $ p < 0.01$ compared with the best baseline results.
The `-' symbol denotes that the corresponding item is not reported in the original paper.
Furthermore, the term 'KG' indicates the model is augmented with a knowledge graph.
    }
    \vspace{-3mm}
\hskip -7pt
  \resizebox{1.01\textwidth}{!}{%
     \def\narrtablewidth{0.77 cm}
     \def\maintablewidth{0.77 cm}
     \def\widetablewidth{0.94 cm}
     \def\maxtablewidth{0.82 cm}
     \def\firsttablewidth{0.93 cm}

    \begin{tabular}{
    p{\firsttablewidth} wl{2.23 cm} wl{1.53 cm}
    p{\narrtablewidth} p{\narrtablewidth}
    p{\narrtablewidth} p{\narrtablewidth} 
    p{\narrtablewidth} p{\narrtablewidth} 
    p{\narrtablewidth} p{\widetablewidth}
     p{\narrtablewidth} p{\narrtablewidth}
    p{\narrtablewidth} p{\narrtablewidth} 
    p{\narrtablewidth} p{\narrtablewidth}
    p{\widetablewidth} p{\narrtablewidth} 
    }
    \hline
     & & & \multicolumn{8}{c}{\textbf{IEMOCAP}} & \multicolumn{7}{c}{\textbf{MELD}} \\
       \cmidrule[1pt](r){4-11}\cmidrule[1pt](r){12-18}
     Input & Model & Embedding & Hap & Sad & Neut & Ang & Exci & Frus & Acc & W-F1 & Neut & Surp & Sad & Joy  & Ang & Acc & W-F1 \\
\hline
\multirow{9}{*}{Text} & DiaGCN~\cite{ghosal-dgcn-2019} & Glove & 51.57 & 80.48 & 57.69 & 53.95 & 72.81 & 57.33 & 63.22 & 62.89 & 75.97 & 46.05 & 19.60 & 51.20 & 40.83 & 58.62 & 56.36 \\
 & HiTrans~\cite{lihitrans-coling-2020} & Bert-Base & - & - & - & - & - & - & - & 64.50 & - & - & - & - & - & - &  61.94 \\
 & RGAT~\cite{ishrga-emnlp-2020}  & Bert-Base& - & - & - & - &  - & - & - & 65.22 & - & - & - & - & - & - &  60.91 \\
 & TUCORE~\cite{leechoi-emnlp-2021} & Robe-Large& - & -  & - &  - & - & -  & -  & -  & -  & -  & - & -  & -  & - & 65.36 \\
 & DiaCRN~\cite{hu-dcrn-2021} & Glove & 53.23 & 83.37 & 62.96 & 66.09 & 75.4 & 66.07 & 67.16 & 67.21 & 77.01 & 50.10 & 26.63 & 52.77 & 45.15 &  61.11 & 58.67 \\
 & EmoFlow~\cite{song-flow-2022} & Robe-Large & - & - & - & - & - & - &  -& - & -  & - & - & - & - & -  &  66.50 \\
\hline
\multirow{4}{*}{T+KG}
& TODKAT~\cite{zhu-topic-2021} & Robe-Large & - & - & - & - & -  & - & - & 61.33 & - & - & - & - & - & - & 65.47 \\
& CoMpM ~\cite{leecmw-naacl-2022} & Robe-Large & - & - & - & - & - & - & - & 69.46 & - & - & - & - & - & - &  66.52 \\
& COMSMIC~\cite{ghosal-cosmic-2020} & Robe-Large & -& -  & -  & - & - & - & - & 65.28 & - & - & - & - & - & - & 65.21 \\
& SKSEC~\cite{gensec-affc-2022} & Robe-Large & - & -  &  - & -  &  - &  - &  - & 66.47 & - & - & - & - & - & - & 66.21 \\
\hline
\multirow{9}{*}{MM}
& TFN~\cite{zadehtf-emnlp-2017} & Glove & 37.26 & 65.21 & 51.03 & 54.64 & 58.75 & 56.98 & 55.02 & 55.13 & 77.43 & 47.89 & 18.06 & 51.28 & 44.15 &  60.77 & 57.74 \\
& MFN~\cite{zadehmf-aaai-2018} & Glove & 48.19 & 73.41 & 56.28 & 63.04 & 64.11 & 61.82 & 61.24 & 61.60 & 77.27 & 48.29 & 23.24 & 52.63 & 41.32 & 60.80 & 57.80 \\
& DiaRNN~\cite{majaar-aaai-2019} & Word2vec & 32.2 & 80.26 & 57.89 & 62.82 & 73.87 & 59.76 & 63.52 & 62.89 & 76.97 & 47.69 & 20.41 & 50.92 & 45.52 & 60.31 & 57.66 \\
& MMGCN~\cite{humm-acl-2021} & FastText& 45.14 & 77.16 & 66.36 & 68.82 & 74.71 & 61.04 & 66.36 &
66.26 & 76.33 & 48.15 & 26.74 & 53.02 & 46.09 & 60.42 & 58.31 \\
& MetaDrop~\cite{chelw-mm-2021} & Robe-Large & - & - & - & - & - & - & 69.38 & 69.59 &  - & - & -  & - & - &  66.63 & 66.30 \\
& DiaTRM~\cite{maoeme-emnlp-2021} & Bert-Base & - & - & - & - & - & - & 69.50 & 69.70 & - &  - & - & - & - & 65.70 & 63.50 \\
& MM-DFN~\cite{humfn-icassp-2022} & FastText& 42.22 & 78.98 & 66.42 & \textbf{69.77} & 75.56 & 66.33 & 68.21 & 68.18 & 77.76 & 50.69 & 22.93 & 54.78 & 47.82  & 62.49 & 59.46 \\
& M2FNet~\cite{chumfn-cvprw-2022} & Robe-Large$_e$ & \textbf{60.00} & 82.11 & 65.88 & 68.21 & 72.6 & \textbf{68.31} & 69.86 & 69.69 & 80.06 & 58.66 & \textbf{47.03} & 65.5 & 55.25 &  \uwave{67.85} & \uwave{66.71} \\
& UniMSE~\cite{hutum-emnlp-2022} & T5-Base & - & - & - & - & - & - & \uwave{70.56} & \uwave{70.66} & - & - & - & - & & 65.09 & 65.51 \\
\hline
Our & DF-ERC
& Robe-Large & 56.37 & \textbf{84.36} & \textbf{71.13} & 67.46 & \textbf{79.11} & 66.23 & \textbf{71.84}$^\star$ & \textbf{71.75}$^\star$ & \textbf{80.17} & \textbf{60.27} & 43.89 & \textbf{65.93} & 55.50 & \textbf{68.28}$^\star$ & \textbf{67.03}$^\star$ \\
   \hline
    \end{tabular}
    }
    \label{tab:main}
\end{table*}

\subsection{Prediction and Learning}
Then, the fused representation $\bm{h}_i^e$ is used for emotion recognition, which is performed as follows:
\begin{gather}
\bm{y}_i = \mathrm{Softmax}(\mathbf{MLP}^c(\bm{h}^e_i)),
\end{gather}
where $\bm{y}_i$ represents the probability of predicted emotion.
We use the cross-entropy loss function for training, which is defined as follows:
\begin{gather}
\mathcal{L}_{emo} = -\sum_{i=1}^n\log{\bm{y}_{i,\mathrm{I}^*_i}},
\end{gather}
where $\mathrm{I}^*_i$ denotes the  golden label index of the utterance $u_i$.
During the learning stage, our training loss functions consist of the following parts:
\begin{gather}
\label{eq:loss}
\mathcal{L}=\alpha_1\mathcal{L}_{cl}+\alpha_2\mathcal{L}_{con}+\alpha_3\mathcal{L}_{sim} + \mathcal{L}_{emo},
\end{gather}
where $\alpha_{1 \textrm{--} 3} \in (0,1]$ are hyper-parameters.

\section{Experiments75}

\subsection{Implementation Details}

\paratitle{Datasets.}
We conducted experiments using two publicly available MM-ERC datasets, namely MELD~\cite{poriaHMNCM-acl-2019} and IEMOCAP~\cite{busied-lre-2007}.
Both of which include text, audio, and video modalities. 
The data split used in our experiments follows previous work~\cite{humm-acl-2021, humfn-icassp-2022}, and the detailed corpus statistics are presented in Section~\ref{sec:data} of the Appendix.

\paratitle{Settings.}
Following previous work~\cite{song-flow-2022,chumfn-cvprw-2022, chelw-mm-2021}, we adopt RoBERTa-large~\cite{liurop-arxiv-2019} as our PLM encoder, with a hidden state dimension of 1024. 
For the MELD dataset, we empirically set hyperparameters $\alpha_{1-3}$ to 0.3, 0.8, and 0.3, respectively. 
For the IEMOCAP dataset, we set the values to 0.2, 0.9, and 1.0. 
More details about the hyperparameter settings can be found in Section~\ref{app:setting} of the Appendix.
We determine all hyperparameters through development experiments on the validation sets.
We report the final results as an average over five random seeds, and we consider the evaluation score significant when the p-value is less than 0.01.

\paratitle{Baselines.}
We compare the performance of our DF-ERC model with several strong baselines, including text-based models, text + knowledge-enhanced models, and multimodal-based models, which are listed in Table~\ref{tab:main}.
We also present the pre-trained embedding weight of each model, most of which use Robert-Large to encode the text content.
We present the original results for each baseline, except TODKAT~\cite{zhu-topic-2021} where we use updated results from the paper's repository.\footnote{\url{https://github.com/something678/TodKat}}.
The training datasets in UniMSE~\cite{hutum-emnlp-2022} are merged from three corpora, possibly contributing to improved performance.

\subsection{Main Results}

Table~\ref{tab:main} presents the experimental results for two benchmark datasets.
When compared to text-based models, DF-ERC significantly improves performance scores. Specifically, for the MELD dataset, DF-ERC improves accuracy (Acc) and weighted F1 (W-F1) scores by 7.17 and 0.53 percentage points (hereafter, `points') as compared to the best-performing baseline, respectively.
Similarly, for the IEMOCAP dataset, DF-ERC improves Acc and W-F1 scores by 4.68 and 4.54 points, respectively.
These findings underscore the efficacy of incorporating multimodal information and the efficient integration of multimodal features.

Interestingly, without utilizing any external knowledge, DF-ERC still surpasses models that rely on external knowledge, such as TODKAT, CoMpM, COMSMIC, and SKSEC.
The table demonstrates that DF-ERC improves the W-F1 scores by 0.51 and 2.29 points for the MELD and IEMOCAP datasets, respectively.
These findings suggest that multimodal information effectively compensates for the absence of external knowledge, thus enhancing emotion recognition performance.

Additionally, DF-ERC achieves the best performance among all multimodal-based models.
On the MELD dataset, DF-ERC improves Acc and W-F1 scores by 0.43 and 0.32 points, respectively.
On the IEMOCAP dataset, the improvements are 1.28 and 1.09 points.
These results indicate that DF-ERC is adept at discerning the differences and weighted contributions of each type of multimodal feature.
Consequently, DF-ERC optimally utilizes multimodal features to bolster multimodal emotion recognition in conversation.

Lastly, we present the performance scores for each type of emotion, revealing that DF-ERC achieves the best performance for most emotions, thereby demonstrating the robustness of our model.
It also contributes to the superior overall performance of our model.

\begin{table}[!t]
    \centering
    \setlength{\tabcolsep}{3pt} 
     \def\narrtablewidth{1.45 cm}
     \def\firsttablewidth{1.55 cm}
    \caption{Ablation studies for DDM, CFM, and CRM, 
    where '+Att' denotes the application of a self-attention mechanism for feature fusion, 
    where `full' and `zero' means the weight of contextual features (Eq.~\eqref{eqn:context_feature}), i.e., using full contextual features or none of them.
    The notions `Utterance' and `Modality' correspond to the removal of utterance-level and modality-level contrastive learning, respectively.
    }
    \begin{tabular}
    {
    p{\firsttablewidth} p{\narrtablewidth} p{\narrtablewidth} p{\narrtablewidth} p{\narrtablewidth} 
    }
    \hline
        \multirow{2}{*}{ Model }& \multicolumn{2}{c}{\textbf{MELD}} & \multicolumn{2}{c}{\textbf{IEMOCAP}} \\
        \cmidrule(r){2-3} \cmidrule(r){4-5} 
        & Acc & W-F1 & Acc & W-F1 \\ \hline
         DF-ERC & 68.28 & 67.03 & 71.84 & 71.75 \\
        \hdashline
- DDM & 66.36\mfnt{($\downarrow$ 1.92)} & 65.59\mfnt{($\downarrow$ 1.44)} & 69.99\mfnt{($\downarrow$ 1.85)} & 69.81\mfnt{($\downarrow$ 1.94)} \\
- Utterance & 66.91\mfnt{($\downarrow$ 1.37)} & 65.92\mfnt{($\downarrow$ 1.11)} &71.45\mfnt{($\downarrow$ 0.39)} & 70.55\mfnt{($\downarrow$ 1.20)} \\
- Modality & 67.78\mfnt{($\downarrow$ 0.50)} & 66.48\mfnt{($\downarrow$ 0.55)} & 70.95\mfnt{($\downarrow$ 0.89)} &	70.94\mfnt{($\downarrow$ 0.81)} \\
 \hline
 - CFM & 66.51\mfnt{($\downarrow$ 1.77)} & 65.49\mfnt{($\downarrow$ 1.54)} & 69.69\mfnt{($\downarrow$ 2.15)} & 69.56\mfnt{($\downarrow$ 2.19)} \\
 \quad + Att & 65.63\mfnt{($\downarrow$ 2.65)} & 65.87\mfnt{($\downarrow$ 1.16)} & 71.41\mfnt{($\downarrow$ 0.43)} & 71.40\mfnt{($\downarrow$ 0.35)}  \\
 \hline
- CRM(full) & 64.98\mfnt{($\downarrow$ 3.30)} & 65.07\mfnt{($\downarrow$ 1.96)} & 69.56\mfnt{($\downarrow$ 2.28)} & 69.42\mfnt{($\downarrow$ 2.33)} \\
- CRM(zero) & 65.06\mfnt{($\downarrow$ 3.22)} & 65.09\mfnt{($\downarrow$ 1.94)} & 69.75\mfnt{($\downarrow$ 2.09)} & 69.71\mfnt{($\downarrow$ 2.04)} \\
\quad + Att &  66.70\mfnt{($\downarrow$ 1.58)} & 65.97\mfnt{($\downarrow$ 1.06)} & 70.06\mfnt{($\downarrow$ 1.78)} &	69.34\mfnt{($\downarrow$ 2.41)} \\
        \hline
    \end{tabular}
    \label{tab:ablation0}
\end{table}

\begin{table}[!t]
    \centering
    \setlength{\tabcolsep}{3pt} 
     \def\narrtablewidth{1.21 cm}
     \def\firsttablewidth{2.30 cm}
    \caption{Ablation studies for different modalities, where T, A, and V denote Text, Audio, and Video, respectively.}
    \vspace{-1mm}
    \begin{tabular}
    {
    p{\firsttablewidth} 
    p{\narrtablewidth} p{\narrtablewidth} p{\narrtablewidth} p{\narrtablewidth} 
    }
    \hline
        \multirow{2}{*}{ Model }& \multicolumn{2}{c}{\textbf{MELD}} & \multicolumn{2}{c}{\textbf{IEMOCAP}} \\
        \cmidrule(r){2-3} \cmidrule(r){4-5} 
        & Acc & W-F1 & Acc & W-F1 \\ \hline
         DF-ERC(T+A+V) & 68.28 & 67.03  & 71.84 & 71.75  \\ 
        \hdashline
        T & 65.17 & 64.54 & 65.13 & 65.46 \\
         A & 43.83 & 41.72 & 41.47 & 38.62 \\
         V & 46.05 & 36.65 & 32.84 & 22.70 \\
        T+V & 65.33 & 64.54 & 70.61 & 69.49 \\
         T+V & 65.10 & 64.95 & 66.17 & 65.89 \\
         A+V & 48.70 & 45.00 & 55.70 & 55.07 \\
         \hline
    \end{tabular}
    \label{tab:ablation1}
\end{table}

\subsection{Ablation Studies}
\vspace{-2mm}
\paratitle{Ablation studies for DDM, CFM, and CRM.}
As shown in Table~\ref{tab:ablation0}, 
we observe that upon the removal of the DDM, utterance-level disentanglement, and modality-level disentanglement, there is a decrease in the model's performance on both datasets, indicating feature disentanglement is crucial for emotion prediction.
Additionally, the model's performance drops by around one point without the CFM.
While an attention-based mechanism does offer some assistance, it remains inadequate when compared with the CFM.
We attribute this to the fact that different modalities make varying contributions, and the use of a contribution-aware approach allows for the adaptive learning of weights for different modality features, resulting in a better fusion of multimodal features.

Last but not least, CRM also provides assistance in emotion recognition.
After removing the CRM module, we directly set the context weight to 1 (full weight) or 0 (zero weight), signifying the introduction of all or no context information.
We observe that these static weights impair the model's performance, resulting in a drop of more than 2 points on both the MELD and IEMOCAP datasets.
This finding suggests that context representations are not inherently useful or useless.
We ultimately achieve better performance using the CRM to determine the weights of the context.

\begin{figure}[!t]
\includegraphics[width=0.98\columnwidth]{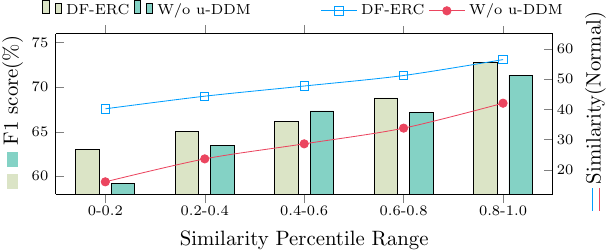}
\caption{
Influence of utterance-level disentanglement (u-DDM) on similarities between modalities and final performance.
The x-axis denotes the similarity percentile range, dividing utterances into five groups based on their corresponding similarity values(see $\psi_i$ in Eq.~\eqref{eq:phi}).
}
\label{fig:uddm}
\end{figure}

\paratitle{Ablation studies for modalities.}
As demonstrated in Table~\ref{tab:ablation1}, our initial findings reveal that the text-based model outperforms other modality-based models, providing evidence of the dominance of text as a modality, which is consistent with previous research findings~\cite{hazarika-mm-2020}.
However, compared to state-of-the-art text-based models in Table~\ref{tab:main}, our text-based model exhibits slightly lower performance.
This is because DF-ERC primarily focuses on multimodal inputs and lacks complex structures specifically tailored for unimodal inputs, which slightly limits its performance.
Nonetheless, incorporating audio and video features into the model with our efficient fusion techniques (i.e., CFM and CRM) leads to a significant improvement in performance.

\subsection{In-depth Analysis}
\label{sec:indep}
To further investigate the effectiveness of DF-ERC, we conduct in-depth analyses to answer the following questions:

\paratitle{Q1: How does utterance-level feature disentanglement influence the feature fusion process?}
We analyzed the influence of utterance-level disentanglement on context weight and final performance.
As shown in Figure~\ref{fig:uddm}, we divided all instances into five groups based on the similarities between modalities, sorted in ascending order.
From the similarity curve, we found that the use of utterance-level disentanglement can significantly improve the similarities between modalities within an utterance, demonstrating that it effectively captures utterance-level information.
Furthermore, we observed that utterance-level disentanglement is more effective in the case of utterances with a lower similarity between modalities (demonstrated by the larger F1 score gap).
This is because low similarity often indicates that the features of the utterance are not fully exploited, and adding utterance-level disentanglement brings the utterance-level distance closer, thus improving similarity to a greater extent.
Finally, considering the F1 metric, utterance-level disentanglement contributes more to the performance of utterances with low similarities, indicating that it can improve the final performance by leveraging utterance-level similarity.

\begin{figure}[!t]
\includegraphics[width=0.98\columnwidth]{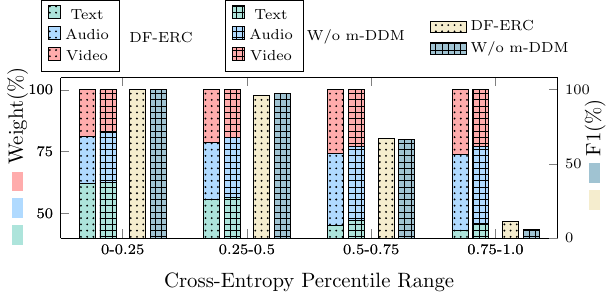}
\caption{
The influence of modality-level disentanglement (m-DDM) on the weight of different modalities and the overall performance. 
The x-axis represents the utterances sorted by their cross-entropy values in ascending order and divided into four groups based on percentiles (25th, 50th, and 75th percentiles). 
The dual y-axis shows the average weight of each modality within each group (left) and the corresponding F1 score (right). 
The equation for calculating the weight of each modality within an utterance is given by Eq.~\eqref{eq:omega}.
}
\label{fig:mddm}
\end{figure}

\paratitle{Q2: How does modality-level feature disentanglement influence the feature fusion process?}
We analyzed the effect of modality-level feature disentanglement on the weight of each modality and the final F1 score. 
From the curves depicted in Figure~\ref{fig:mddm}, it is evident that integrating modality disentanglement resulted in increased weights for the video and audio modalities, especially in utterances with higher cross-entropy values.
This can be attributed to the fact that modality disentanglement enables each modality's unique characteristics to be fully exploited, resulting in a subtle enhancement of weaker modalities, such as video and audio. 
Moreover, this increase of weight for weaker modalities is more evident in utterances with suboptimal prediction results, where the text modality does not perform well. 
We observed the most substantial improvement in the F1 score for utterances with poorer prediction results, approximately 5 points improvement for those with cross-entropy ranking percentile > 0.75.
This suggests that the integration of modality-level feature disentanglement can lead to more accurate predictions in challenging situations.

\begin{figure}[!t]
\centering
\subfloat[MELD]{{\includegraphics[width=0.48\columnwidth]{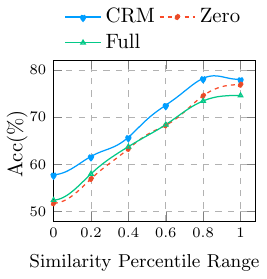} }}%
\subfloat[IEMOCAP]{{\includegraphics[width=0.48\columnwidth]{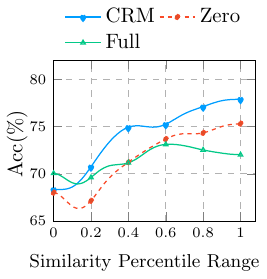} }}%

\caption{
Correlation between performance and modality consistency, where a larger value of the x-axis represents a higher modality consistency.
We use similarity to represent consistency, where a higher similarity of different modalities indicates a higher consistency.
CRM: Our proposed Context Refusion Mechanism dynamically determines how much contextual features are used based on modality consistency;
Zero: using no contextual features;
Full: using 100\% contextual features.
}
\label{fig:exp_sim}%
\end{figure}

\paratitle{Q3: Do the CRM context weights decided by modality consistency really work?}
To verify the effectiveness of the modality similarity comparison module, we study the relationship between prediction performance and modality similarity. 
Additionally, we include the performance under full weight (incorporating all context representations) and zero weight (excluding all context representations). 
Figure~\ref{fig:exp_sim} displays these results, with the x-axis representing the instance's similarity score ranking among all instances, while larger x values denote higher similarity. 
Firstly, we find that as modality similarity improves, the overall performance of the model also increases gradually. 
This is because the more similar the three modalities are, the more consistency they exhibit in emotion recognition, resulting in higher prediction scores. 
Secondly, when comparing full weight to zero weight, we observe that for utterances with relatively small similarities, the performance of full weight is superior to that of zero weight. 
This is because when the discrepancy between different modalities within an utterance is substantial, introducing context utterance representations can help to better recognize emotions. 
Conversely, as similarity increases and the discrepancy between different modalities diminishes, context representations interfere with emotional judgments. 
Thus, for the latter half of the figures, the performance score of zero weight outperforms that of full weight. 
Finally, regardless of whether full weight or zero weight is used, both approaches have limitations in that they can not be flexibly adjusted as the consistency of modalities. 
Our CRM can adjust context weight according to the consistency between modalities, achieving the best performance in all instances and verifying that DF-ERC effectively captures the relationship between multimodal features and context features.

\section{Conclusion}
In this work, we introduce a novel MM-ERC system that emphasizes both feature disentanglement and fusion while taking into account both multimodalities and conversational contexts.
Our proposed Dual-level Disentanglement Mechanism (DDM) successfully disentangles modality- and utterance-level features using contrastive learning, while the Contribution-aware Fusion Mechanism (CFM) and Context Refusion Mechanism (CRM) fuse multimodal and contextual features effectively.
Extensive experiments on two public datasets demonstrate that DF-ERC achieves the best performance compared with 19 models. 
Ablation studies and in-depth analyses substantiate the rationality of our approaches for controllable fusing multimodal and context features. 
Intuitively, our proposed approaches are not limited to emotion recognition in dialogs, and we will evaluate them on other multimodal tasks in the future.

\begin{acks}
This work is supported by the National Key Research and Development Program of China (Grant No. 2022YFB3103602, Grant No. 2017YFC1200500), the National Natural Science Foundation of China (Grant No. 62176187), China Scholarship Council (CSC), NExT Research Center, and the Research Foundation of Ministry of Education of China (Grant No. 18JZD015).
\end{acks}

\clearpage

\bibliographystyle{ACM-Reference-Format}
\balance
\bibliography{acmmm}

\clearpage
\appendix

\section{Dataset \& Experiment Details}
\label{app:baseline}

\subsection{Dataset}
\label{sec:data}
We provide detailed descriptions of the two datasets used in this study: MELD and IEMOCAP.

\paratitle{MELD.} MELD is a multi-party dialogue dataset consisting of conversation snippets from the TV show \textit{Friends}.
The dataset includes 1,433 dialogues and 13,708 utterances, with an average of 9.6 turns per dialogue, and features 378 unique speakers.
Each utterance in the dataset is labeled with one of seven emotions, namely \textit{joy}, \textit{sadness}, \textit{neutral}, \textit{surprise}, \textit{anger}, \textit{fear}, and \textit{disgust}, based on the emotion conveyed by the speaker.
The detailed statistics for the MELD dataset are provided in Table~\ref{tab:data0}.

\begin{table}[h]
\fontsize{9.9}{11}\selectfont
    \caption{ 
    The statistics for the MELD dataset. 
    In the dataset, there are seven different types of emotions: Neutral, Surprise, Fear, Sadness, Joy, Disgust, and Anger.
    }
\resizebox{1.00\linewidth}{!}{
     \def\narrtablewidth{0.67 cm}
     \def\maintablewidth{0.77 cm}
     \def\widetablewidth{0.82 cm}
     \def\maxtablewidth{0.82 cm}
    \begin{tabular}
    {wl{1.13 cm} 
    p{\narrtablewidth} p{\narrtablewidth} p{\narrtablewidth} 
    p{\narrtablewidth} p{\narrtablewidth} p{\narrtablewidth} 
    p{\narrtablewidth} p{\narrtablewidth} 
    }
    \hline
      & \textbf{Neut} & \textbf{Surp} & \textbf{Fea} & \textbf{Sad} & \textbf{Joy} & \textbf{Dis} & \textbf{Ang} & \textbf{Total} \\
    \hline
    Train & 4,710 & 1,205 & 268 & 271 & 683 & 1,743 & 1,109 & 9,989 \\
    Valid & 470 & 150 & 40& 22 & 111 & 163 & 153 & 1,109  \\
    Test & 1,256 & 281 & 50 & 68 & 208 & 402 & 345 & 2,610 \\
    \hdashline
    Total & 6,436 & 1,636 & 358 & 361 & 1,002 & 2,308 & 1,607 & 13,708 \\
    \hline
    \end{tabular}
    }
    \label{tab:data0}
\end{table}

\paratitle{IEMOCAP.}
IEMOCAP is another dataset used in this study and comprises a total of 151 dialogues and 7,433 utterances.
The dataset features two speakers interacting in each session, with a total of 10 speakers across all dialogues.
Each utterance in the dataset is labeled with one of six emotions: \textit{happy}, \textit{sad}, \textit{neutral}, \textit{angry}, \textit{excited}, or \textit{frustrated}.
The detailed statistics for the IEMOCAP dataset are provided in Table~\ref{tab:data1}.

\begin{table}[ht]
\fontsize{9.9}{11}\selectfont
    \caption{ 
    The statistics for the IEMOCAP dataset. 
    In the dataset, there are six different types of emotions: \textbf{Happy}, \textbf{Sad}, \textbf{Neutral}, \textbf{Angry}, \textbf{Excited}, and \textbf{Frustrated}. 
    }
\resizebox{1.00\linewidth}{!}{
     \def\narrtablewidth{0.67 cm}
     \def\maintablewidth{0.77 cm}
     \def\widetablewidth{0.82 cm}
     \def\maxtablewidth{0.82 cm}
    \begin{tabular}
    {wl{1.13 cm} 
    p{\narrtablewidth} p{\narrtablewidth} p{\narrtablewidth} 
    p{\narrtablewidth} p{\narrtablewidth} p{\narrtablewidth} 
    p{\narrtablewidth} p{\narrtablewidth} 
    }
    \hline
      & \textbf{Hap} & \textbf{Sad} & \textbf{Neut} & \textbf{Ang} & \textbf{Exci} & \textbf{Frus} & \textbf{Total} \\
    \hline
    Train & 448 & 736 & 1,229 & 834 & 653 & 1,346 & 5,246 \\
    Valid & 56 & 103 & 95 & 99 & 89 & 122 & 564 \\
    Test & 144 & 245 & 384 & 170 & 299 & 381 & 1,623 \\
    \hdashline
    Total & 648 & 1,084 & 1,708 & 1,103 & 1,041 & 1,849 & 7,433 \\
    \hline
    \end{tabular}
    }
    \label{tab:data1}
\end{table}

\subsection{Settings}
\label{app:setting}
We employ RoBERTa-Large to encode text content without any additional pre-processing operations. 
We utilize the AdamW~\cite{loshdwd-iclr-2019} optimizer and LR scheduler with a warm-up mechanism for parameter optimization.
The learning rates for the PLM layer and non-PLM layer are set to 1e-5 and 1e-3, respectively.
Utterances exceeding 256 tokens are clipped to meet the model's input length requirement and reduce memory usage.
Furthermore, we apply a dropout layer with a rate of 0.2 after the encoder to further enhance the performance of our model.
We set the batch size to 8 and 4 for the MELD and IEMOCAP datasets, respectively.
We set the temperature in DDM to 0.5 for Eq.~\eqref{eq:clm} and 0.3 for Eq.~\eqref{eq:clu}.
All experiments are conducted on Ubuntu systems with two RTX A5000 GPUs.
Additional parameters are shown in Table \ref{tab:apphyp}.

\section{Additional Experiment}
In this section, we present more experiment results to investigate the performance of DF-ERC.

\begin{table}[!h]
\setlength{\tabcolsep}{3.mm}
    \caption{
    Hyperparameters setting.
    }\label{param-del}
\vspace{-2mm}
    \centering
    \begin{tabular}{llll}
    \hline
    Parameter/Module & MELD & IEMOCAP \\ 
    \hline
    Encoder & \\
        \quad Text Embedding Dim.& 1024 & 1024 \\
        \quad Audio Embedding Dim.& 300 & 1582 \\
        \quad Video Embedding Dim. & 342 & 342 \\
    \hline
    DDM & \\
        \quad MLP$^m_{t/a/v}$ Output Dim. (Eq. 4) & 300 & 300 \\
        \quad MLP$^u_{t/a/v}$ Output Dim. (Eq. 6) & 300 & 300 \\
    \hline
    CFM &   \\
        \quad MLP$^r_{t/a/v}$ Output Dim. (Eq. 19) & 500 & 500 \\
    \hline
    CRM & \\
        \quad BiLSTM Hidden Dim. (Eq. 23) & 300 & 300 \\
        \quad $\beta$ (Eq. 21) & 0.1 & 0.1 \\
        \hline
    Training & \\
        \quad Epoch size & 10 & 10 \\
        \quad Max grad norm & 1.0 & 1.0 \\
        \quad Warmup steps &  100 & 100 \\
        \quad Weight Decay &  0.01 & 0.01 \\
    \cdashline{1-4}
    \hline
    \end{tabular}
\label{tab:apphyp}
\end{table}

\begin{figure}[!h]
\includegraphics[width=0.98\columnwidth]{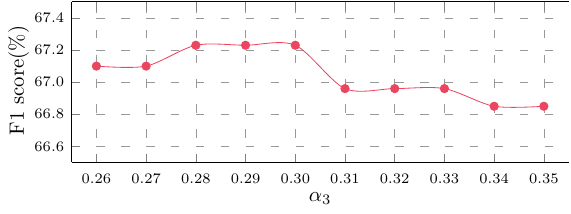}
\caption{Trend in F1 score with respect to the changes in $\alpha_3$ in Eq.~\eqref{eq:loss}.}
\label{fig:alpha}
\end{figure}

\paratitle{Hyper-parameter Analysis}
In our study, we introduce certain hyperparameters and adopt a grid search strategy for their optimization.
An example of these parameters is $\alpha_3$, as referenced in Eq.\eqref{eq:loss}.
To assess the effect of the parameter and evaluate the robustness of DF-ERC, we document the variation in the F1 score as $\alpha_3$ is adjusted within a particular range.
As depicted in Figure~\ref{fig:alpha}, the F1 score fluctuates slightly with changes in $\alpha_3$, peaking when $\alpha_3=0.3$. 
Performance experiences a minor decline when $\alpha_3$ deviates from this optimal value, illustrating the robustness of DF-ERC around $\alpha_3=0.3$.

\begin{figure*}[!t]
\centering
\subfloat[Modality-MELD]{{\includegraphics[width=0.48\columnwidth]{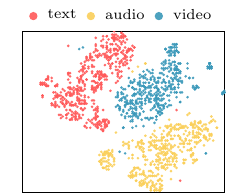} }}%
\subfloat[Utterance-MELD]{{\includegraphics[width=0.48\columnwidth]{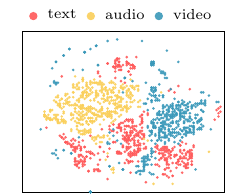} }}%
\subfloat[Modality-IEMOCAP]{{\includegraphics[width=0.48\columnwidth]{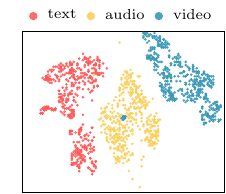} }}%
\subfloat[Utterance-IEMOCAP]{{\includegraphics[width=0.48\columnwidth]{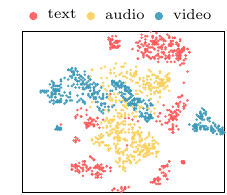} }}%
\vspace{-1mm}
\caption{
T-SNE~\cite{matvdu-JMLR-2008} visualization of multimodal features after applying DDM in modality and utterance levels. 
}
\label{fig:exp_cl}%
\vspace{-5mm}
\end{figure*}

\paratitle{Visualization for feature disentanglement.}
To visualize the effectiveness of DDM for feature disentanglement, we analyze the distribution of the three modalities after modality-level disentanglement (see Eq.~\eqref{eq:tmi}) and utterance-level disentanglement (see Eq.~\eqref{eq:tui}) using t-SNE~\cite{matvdu-JMLR-2008}, as shown in Figure~\ref{fig:exp_cl}.
The result indicates that modality-level contrastive learning effectively disentangles the three modalities from each other.
Furthermore, we also observe that utterance-level disentanglement can align features within an utterance by entangling features from three distinct modality spaces.
These findings highlight the effectiveness of our DDM in controlling the modality distribution in feature space based on the corresponding optimization objective and thus can further improve the emotion recognition performance.

\begin{figure}[!t]
\centering
\subfloat[MELD]{{\includegraphics[width=0.48\columnwidth]{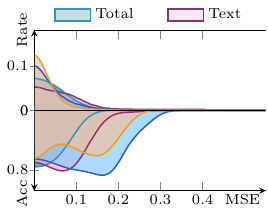} }}%
\subfloat[IEMOCAP]{{\includegraphics[width=0.48\columnwidth]{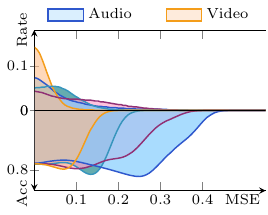} }}%
\vspace{-3mm}
\caption{
Virtualization of the correlation between prediction performance and MSE (Eq.~\eqref{eq:mseomega}, the difference between TCP and modality contribution) in CFM.
X-axis: MSE; Upper Y-axis: Proportions of MSE; Lower Y-axis: Predicted accuracy using each modality.
}
\vspace{-2mm}
\label{fig:exp_mse}%
\end{figure}

\paratitle{The contribution of CFM for final performance.}
To verify the effect of CFM, we investigated the relationship between the prediction effect (evaluated using MSE) of TCP and the final emotion recognition performance.
Figure~\ref{fig:exp_mse} shows that for the majority of utterances, the MSE of the TCP prediction is less than 0.1, indicating satisfactory performance for TCP.
However, it should be noted that a better TCP prediction for each modality does not necessarily result in a higher emotion prediction score, as shown by the wave pattern in the bottom half of the figure, which suggests that TCP has some limitations as a modality contribution evaluator.
Nonetheless, after averaging the prediction MSE of the three modalities within an utterance, we observed a gradual increase in performance with a smoother tendency as the MSE decreased. 
This demonstrates that, overall, a predicted contribution weight can guide the model to assign the optimal weight for each modality and thus achieve better final performance.

\end{document}